# Human-Robot Mutual Learning through Affective-Linguistic Interaction and Differential Outcomes Training [Pre-Print]

Emilia Heikkinen[1]*, Elsa Silvennoinen[1]*, Imran Khan[1]*, Zakaria Lemhaouri[2], Laura Cohen[2], Lola Cañamero[2], and Robert Lowe[1]

[1] Digitalization, Interaction, Cognition, and Emotion (DICE) Lab, Department of Applied IT, University of Gothenburg, Sweden

[2] ETIS Lab, CY Cergy Paris University, France

Corresponding author: Robert Lowe
robert.lowe@ait.gu.se

*Abstract*— **Owing to the recent success of Large Language Models, Modern A.I has been much focused on linguistic interactions with humans but less focused on non-linguistic forms of communication between man and machine. In the present paper, we test how affective-linguistic communication, in combination with differential outcomes training, affects mutual learning in a human-robot context. Taking inspiration from child-caregiver dynamics, our human-robot interaction setup consists of a (simulated) robot attempting to learn how best to communicate internal, homeostatically-controlled needs; while a human "caregiver" attempts to learn the correct object to satisfy the robot's present communicated need. We studied the effects of i) human training type, and ii) robot reinforcement learning type, to assess mutual learning terminal accuracy and rate of learning (as measured by the average reward achieved by the robot). Our results find mutual learning between a human and a robot is significantly improved with Differential Outcomes Training (DOT) compared to Non-DOT (control) conditions. We find further improvements when the robot uses an exploration-exploitation policy selection, compared to purely exploitation policy selection. These findings have implications for utilizing socially assistive robots (SAR) in therapeutic contexts, e.g. for cognitive interventions, and educational applications.**

## I. Introduction (Heading 1)

### A. Socially Assistive Robots as Affective Interactors

The revolutionary speed at which Large Language (and Foundation) Models are developing and being applied to various use cases has included recent attempts to integrate them into socially assistive robots (SARs). However, human-like interaction has many non-linguistic (verbal and non-verbal) components necessary for seamless communication, which require strong consideration, therefore, for any given Human-Robot Interaction (HRI) use case.

---

* The first three authors contributed equally and are listed as co-first authors.

The increased use of SARs for purposes of motivational [1], social [2], pedagogical [3], and therapeutic [4] assistance necessitates addressing the question of what linguistic and non-linguistic components need to be exploited in both humans and robots for permitting long-term engaging interactions. Such non-linguistic components may consist of expressions of internal (homeostatic) needs and other affective components as they affect internal reward and affect representations of the value of particular interactions. Long-term interactions may consist of digital companions (e.g. Tamagotchi-style social robots) but also multi-session interventions for clinical persons such as elderlies or children with cognitive impairments who would benefit from more engaging interactions to help address the problem of non-adherence or dropout during the intervention.

Indeed, Human-robot collaboration and interaction setups with SAR are gaining traction as a promising approach for facilitating and mitigating cognitive interventions [5] with elderlies [6] and also with children [7]. A purpose for SARs in cognitive therapy is to mitigate the problem of treatment adherence by providing engaging and informative feedback. This entails an instructional role in the interventions (i.e. the robot gives task-related feedback, or instructions). Another predominant use case for healthcare concerns use of a SAR as an engaging and informing partner [8]. However, in order to promote engagement with the SAR as a human-like companion or otherwise an agent worthy of attention and interest, feedback not just from robot to human but also from *human to robot* to tailor and sculpt interactions may be beneficial [6], [9].

### B. Mutual Learning

The term *Mutual Learning* is oft-used in psychology and business. Mutual learning can be defined as "involv[ing] a relationship between two people ... [in which] ... both people take turns to act as the teacher and the student." [10]. Mutual learning is also considered as providing a means for complementary skills and learning approaches to be used and shared in relation to educational settings [11]. A notion of mutual learning has been used much in relation to machine learning with respect to multiple learning agents [12]; however, in relation to human-robot interaction it has been less often referred to (but see [9]). Nevertheless, viewing learning in HRI as a process of mutual instruction with joint teacher-student or instructor-follower roles has been gaining traction in SAR applications [13], [14], [6]. Where human interactors are required to apprehend the learning needs and preferences of others in relation to healthcare or therapy the potential for long-term benefits may increase [15]. Cognitive interventions for clinical participants can also potentially benefit given the weeks-to-months multiple sessions that are typically required. In the case of HRI, if participants feel a greater sense of attachment to SAR interactors and empowerment as a result of not merely having a passive (follower) role during the learning, adherence and completion of therapy may also increase. In the present study, mutual learning is measured as a function of the rate of the learning rate carried out by human-robot during the training task.

### C. Learning via Differential Outcomes Training

Another type of (associative) learning, born out of non-human and human-animal learning traditions using either implicit or explicit rewards, is differential outcomes training [16]. Differential outcomes training (DOT) concerns the learning of stimulus-response-outcome associations where 'outcomes' are typically conceived of as specific rewards differential according to type, amount or some other property. Upon a correct response to a specific stimulus, the individual is rewarded with an outcome unique to the particular stimulus-response pair and the association is strengthened. As an example, a participant, in one learning trial, could be presented on a computer screen an image of a particular symbol (e.g. meaningless word stimulus). Following a short interval, simultaneously presented images of a cookie and a cup are presented. The participant selects one of the two images. Only one is 'correct' (cup) and leads to the reward (social praise). In the next learning trial, a different symbol (stimulus) is presented, cookie and cup are presented thereafter and the correct response is to point to the cookie to obtain the reward of 10 dollars. So here a unique stimulus-response pair leads to a unique (differential) rewarding outcome each time. This type of learning leads to increased accuracy and learning rate in humans [17] with a recent meta-analysis [16] also reporting medium-to-large effect sizes on learning speed and accuracy (DOT 'effects').

DOT is hypothesized to function through exploiting an alternative learning route in the brain for processing affective and reward-based stimuli [18]. A dual learning process [19] has even been modelled in relation to individual [20] and social based DOT [21]. Prior research has empirically investigated the application of differential outcomes training within the context of social interaction [22], with recent work finding DOT effects when extending this to human-robot social interactions; in both cooperative [23] and mutual learning [9] tasks.

*D. Study Aims and Research Questions*

The experiment presented in this paper is conducted at the intersection of human-robot interaction, experimental psychology, and cognitive science. The aim is to investigate whether the differential outcomes training (DOT) procedure, expressed through affective non-verbal and verbal expressions from a robot, can improve mutual learning in a human-robot interactive scenario relative to a control (Non-Differential Outcomes) procedure. This study builds on the preliminary work done by [9] in this area a number of ways. First, we seek to address the lack of statistical power in the preliminary results, through an increased sample size. Secondly, we test an additional (exploration-exploitation) policy for robot action selection, aligning with conventional Q-learning approaches [24]. Thirdly, we align our experiment closer to that of DOT experimental paradigms, by limiting the number of possible actions ("outcomes") that the robot can take, as well as fixing the number of epochs (trials) the human completes: offering further contributions to the DOT empirical literature. Explicitly, our research questions are as follows:

RQ1: How does the Differential Outcomes Training (DOT) procedure affect a human's learning of robot's needs, when compared to non-Differential Outcomes Training (Non-DOT)?

RQ2: How does a robot's utilization of an exploration-exploitation, as opposed to a purely exploitation, policy selection affect mutual learning outcomes?

On this basis, we evaluate how mutual learning is affected by two different learning approaches used by both human and robot. Based on the initial findings from [9] and related work [17], [25] our first hypothesis (H1) is that differential outcomes training (DOT) will significantly improve the mutual learning of (a) the human partner's ability to learn a robot's needs, and (b) the robot's ability to learn which verbal utterances yield need-satisfying rewards, compared to a non-differential outcomes training (Non-DOT) control (i.e. where outcomes are randomly selected presented by Reachy given correct answers and not tied to specific stimuli). Our second hypothesis (H2) is that conditions where the robot pursues an exploration-exploitation policy, rather than purely *exploitation*, will also have significant improvements on mutual learning in these contexts. This second hypothesis is in line with related work in reinforcement learning, leveraging exploration-exploitation policy selection to avoid early convergence to a local (suboptimal) maxima.

This study was partially funded by EUTOPIA (Diarienummer: GU 2023/3314)

## II. ROBOT SET UP AND MODEL

In this section, we discuss the set-up of the simulated robot and environment, the robot's action-selection and learning models (Q-learning), and the different types of gestures the robot uses which serve as the (differential) outcomes from the human actor's point of view. Mutual learning is evaluated here through a language acquisition task. Briefly, Reachy must learn to communicate its internal "need" through verbal babbling. Over time, the human partner, who provides objects to Reachy, must learn which object Reachy is "asking for" through each of its babbling sounds. This interaction is inspired by child-caregiver interactions, where small children learn to communicate to older adults, who simultaneously try to understand what is being communicated. An overview of this human-robot interaction is illustrated in Figure 1 and described in more detail in the subsequent subsections.

### A. Simulated Robot Environment

This experiment was conducted with a simulated version of the Reachy robot (version SDK 0.5.1, henceforth referred to as "Reachy"), developed by Pollen Robotics (Unity implemented, version 2020.3.11f1). The simulated version of Reachy was programmed to replicate the functionality of its physical counterpart. Reachy is capable of expressing both non-verbal cues, such as moving its arms, torso, and antennae (Figure 1), as well as expressing verbal cues through the use of text-to-speech, or playing pre-recorded audio files. In this experiment, Reachy can express one of three babbles—"bada", "paba", and "wada"—to communicate with the human partner. The simulated environment also contains three objects – a teddy, a cookie, and a drink – which the human user is able to interact with by presenting them to Reachy (Figure 1).

### B. Robot Architecture: Internal Needs, Motivations, and Action-Selection

Reachy is endowed with a set of internal, homeostatically-controlled needs, which drive its motivational state and subsequent action. Reachy's goal is to maintain each of these needs at (or above) their respective set points, by selecting actions that correct deviations from set point values. This is in line with previous work [26], [27], [28], which has grounded agent action-selection in the maintenance of homeostasis. This model formally described below.

Reachy has three internal homeostatic needs, **H = {Thirst, Hunger, Curiosity}**, and three possible motivational states, **M = {Drink, Eat, Play}**. Each homeostatic variable $i$ takes the range 0—1, has a starting value of 0.5 ($H_{i,t=0} = 0.5$), and set point value of 0.5 ($H_{i,sp} = 0.5$). Each variable also has a decay value, **T = ($\tau_{thirst}$ = 10, $\tau_{hunger}$ = 12, $\tau_{curiosity}$ = 8)**. These are asymmetrical to avoid anticipatory bias. At each time step $t$, each of the homeostatic variables are updated as follows:

$$H_{i,t} = H_{i,t=0} \, e^{(\Delta t/\tau i)} \quad (1)$$

Where **i ∈ H, τ ∈ T, and Δt** the difference between the current time step $t$ and the last time step ($t_k$) when the homeostatic value $H_i$ was satisfied by an object:

$$\Delta t = t_k - t \quad (2)$$

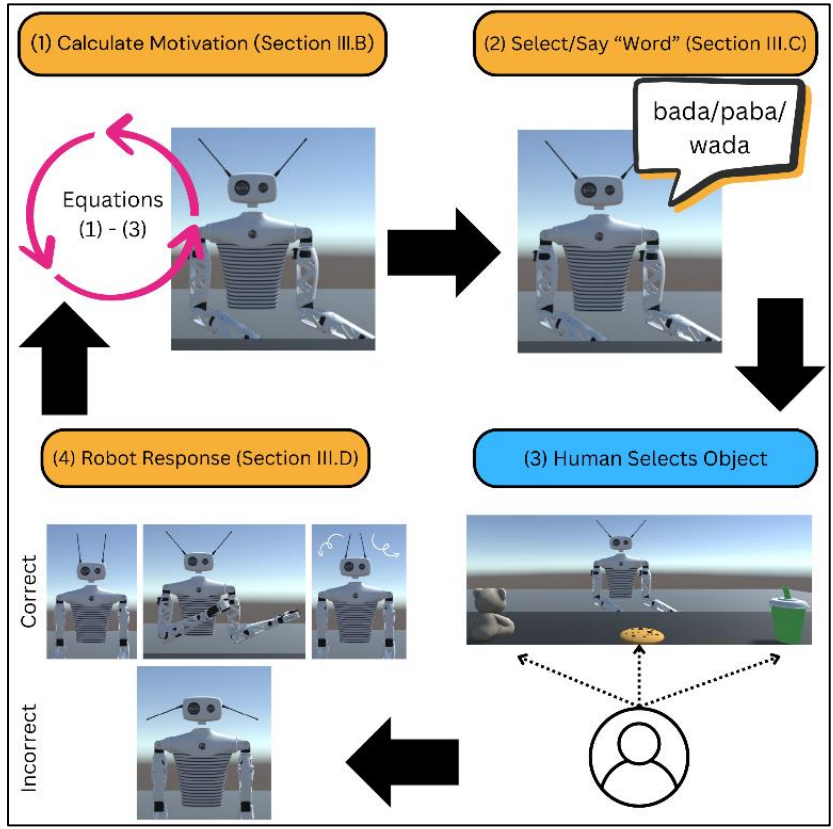

Fig. 1. Depiction of trials of human-robot interaction. (1) a maximum/most urgent homeostatic need ('motivation') among thirst, hunger and curiosity is calculated; (2) the robot expresses this motivation using one of a limited vocabulary of disyllabic sounds ('babbles') initially selected by trial and error. The sound is a "stimulus' from the point of view of the human interactor; (3) the human intuits (initially through trial and error) which of three objects the robot is expressing a need for (toy teddy for curiosity, cookie for hunger, drink for thirst); (4) the robot expresses a differential outcome (a different physical gesture or movement) for each 'correct' object choice, and for the incorrect choice.

The values of each of Reachy's internal needs are used to calculate its motivational states **M**. Reachy can be in one of three motivational states, **M = {Drink, Eat, Play}**, each corresponding to a single homeostatic need. The intensity of each of these three motivations is calculated using the difference of each homeostatic variable $i$ from its set point $H_{i,\text{sp}}$. Thus:

$$M_{s,t} = H_{i,\text{sp}} - H_{i,t} \qquad (3)$$

where **s ∈ M**. The motivational state $s$ returning the largest value is selected as Reachy's motivational state at time $t$. In other words, Reachy's motivational state corresponds to the internal need $H_i$ experiencing the largest deviation from its set point. This motivational state is then used to select an action (a "babble" sound) for Reachy.

*C. Action-Selection Policies & Learning via Q-Learning*

After Reachy's motivational state $s$ has been determined, it selects an action $a$ that it believes will satisfy the respective internal need. Here, an action corresponds to one of three possible babbling sounds that Reachy can make. The babbling sound that Reachy should make is selected (and learned) through Q-learning: a reinforcement learning algorithm that aims to learn optimal actions for a given state in Markovian environments [29]. Each of the possible babbles (actions, *a*) that Reachy can perform for each given (motivational) state *s* is assigned a Q-value, **Q(s,a),** in a Q-table. Reachy selects the action (word) that has the largest value under the selected state. Specifically, for state *s*:

$$a_t = argmax Q_t(a) \quad \text{with probability } 1 - \varepsilon \tag{4}$$

When Reachy utters its chosen babbling, the human partner must infer (by trial and error in early learning trials) which internal need Reachy is communicating and try to select the "correct" object that satisfies the need. Receiving the correct object (i.e. the object that satisfies its current internal need) results in a positive reward (**r = 1**); a negative reward (**r = -1**) otherwise. Reachy aims to learn the optimal associations between its motivational states, *s*, and babbling, *a*, by updating the Q-value through the reward function:

$$Q(s,a) \leftarrow \alpha[Q(s,a)] + r_{o,t} \tag{5}$$

Where $r_{o,t}$ is the reward value **r** associated with the object **o** at time *t*, and $\alpha=.5$; a parameter to avoid q value divergence.

### D. Robot Responses : Differential vs. Non-Differential Outcomes

When Reachy is presented with an object, it performs an affective, audiovisual response to the human. When Reachy is presented with the correct object (i.e. the one that corresponds to its current motivation), it produces a cheerful beeping sound (see Supplementary Materials), and performs one of three types of gestures with its arms, head, or antennae. In line with differential outcomes training (Section I.C), each response-outcome pairing is unique, and so each of the three possible gestures is mapped onto the satisfaction of a specific internal need (see Figure 1). If Reachy is presented with an object that does not correspond to its current motivation, however, it performs a gesture communicating sadness, by looking downwards, lowering its antennae and producing a sad-sounding beeping noise (see Supplementary Materials). These specific responses that Reachy provide have been inspired, in part, by the affective expressions of non-human animals (e.g. such as a dog that lowers its ears when it is sad) and are designed to communicate non-verbal expressions of affective states such as happiness or sadness.

## III. EXPERIMENT DESIGN

### A. Participants

We recruited participants by advertising at the University of Gothenburg campus. In total, we recruited 28 participants (11 identifying as male, 17 identifying as female), between 21 and 66 years of age. 22 out of 28 participants were between the ages of 20-29. Participants could choose to receive information and instructions in either Swedish or English. All participants gave their informed consent before participating in the study. All data collection took place in April 2024. All participant data was anonymized using non-identifying participant IDs.

TABLE I.  OVERVIEW OF 2X2 EXPERIMENTAL DESIGN. DETAILS CAN BE FOUND IN SECTION III.B.

| Condition Names | | Robot Reinforcement Learning Type | |
|---|---|---|---|
| | | Greedy | e-Greedy |
| Human Learning Type | Differential Outcomes | DOT+Greedy | DOT+ε-greedy |
| | Non-Differential Outcomes | Non-DOT+Greedy | Non-DOT+ε-greedy |

*B. Study Design*

This study employs a 2x2 incomplete, within-group design with two independent variables: (1): **Human Training Type:** either differential or non-differential outcome (DOT/Non-DOT) and (2): **Robot Reinforcement Learning Type:** either greedy or ε-greedy (see Table II). Our dependent variables is mean reward over blocks of trials attained by Reachy over the course of each condition, as a measure of mutual learning between Reachy and the human partner (since Reachy reward can only increase if the human is learning to select the correct rewards to satisfy Reachy's needs).

**In relation to (1),** in the **DOT** condition, Reachy performs a unique gesture ("outcome") upon satisfaction of each of its internal needs (i.e. each outcome is unique to a stimulus-response pair, hence "differential" outcomes). In the **Non-DOT** condition, Reachy performs a random gesture upon satisfaction of its internal needs.

**In relation to (2),** in the **greedy** condition, Reachy's action-selection policy is to select the action ("babble") with the largest Q-value (Equation 4). In the **ε-greedy** condition, Reachy takes the action with the largest Q-value with 9 state-action values, and selects a random action with .1 probability ($\varepsilon = 0.1$).

Each condition was run for 25 epochs (trials), with each participant exposed to all four conditions one after another. We dropped the first trial from our results to reduce variance (allow for initial orientation). To control for potential practice effects, the order in which conditions were presented were controlled for (pseudo-randomized) amongst all participants (see Supplementary Materials). Participants were also made aware that Reachy "resets" at the start of each new condition.

We recorded both the mean reward value attained by Reachy over 24 trials (reported as mean values over three blocks of eight for analysis purposes) in each experimental condition. We evaluated: i) *terminal accuracy* – mean reward value for Reachy in the final eight trials (the final block) of each condition. This was in order to evaluate DOT vs Non-DOT performance following early-stage learning; ii) *learning rate* – how performance changed over three blocks of 8 trials (evaluated as mean reward value for Reachy). These evaluations provided our measures of *mutual learning* since Reachy's achieved average reward could not increase unless the human interactor also 'correctly' selected the object that 'rewarded' the robot's dominant present motivation/need. Additionally, we asked a single interview question to the participants ("How did it feel to take part in this task?") at the conclusion of each session.

*C. Experimental Procedure*

Participants were seated in front of the laptop and were briefed on the functionalities of the simulation, how to interact with the simulation, and the objectives of interacting with Reachy (see Supplementary Materials). All data collection was conducted at Lindholmen Campus, University of Gothenburg (Sweden), using the same laptop for all participants. Both researchers were present throughout all experiments. Participants were read the same script at the start of each experiment (see Supplementary Materials) and asked not to interact with the researchers. However, participants were free to take breaks between conditions if they chose. After participants had tested interacting with the simulation, the experiment began. The experimental procedure worked as follows:

At the start of each condition (and each subsequent epoch), Reachy 'experienced' one of three motivational states (Section II.B.) related to each of its needs. To satisfy these needs, Reachy communicates to the human by selecting and vocalizing one of three possible babbles (the "stimulus" to the human, see Figure 1). The role of the human is to infer, from this babbling, which of Reachy's internal needs requires satisfying, and to present the associated object (the "response") to Reachy. This is achieved by pressing one of three keys (c/b/d) on the laptop keyboard. Upon receiving an object, Reachy performs a gesture (the "outcome") depending on whether the object presented is "correct" (i.e. corresponding to its current internal need) or "incorrect" (see Figure 1).

The goal for Reachy is that it must satisfy its own internal homeostatic needs, by learning and expressing the babbling that it believes corresponds to its current internal need. The goal for the human partner is that they must learn the correct object (and, therefore, the internal need) associated with each of Reachy's babbling sounds. Thus, this experiment is set up as a "mutual learning" scenario: both Reachy and the human partner are required to learn different types of associations. Reachy must learn the babbling that best (correctly) communicates its internal need, whereas the human must learn the correct associations between babbling and corresponding object which, in turn, satisfies Reachy's internal need.

## IV. Results

Statistical analysis of the collected data was performed in SPSS. To answer our research questions, we firstly evaluated *terminal accuracy* (assessed with respect to Reachy's attained reward). We conducted a two-way ANOVA with repeated measures to analyse the interaction effect of **Human Training Type** (DOT/Non-DOT) and **Robot Reinforcement Learning Type** (greedy/ε-greedy) on the mean reward attained. We focused this analysis on the final, third, block (i.e. the final eight trials) of each condition. The descriptive statistics of each condition can be found in Table II.

TABLE II. RESULTS FOR MEAN REWARD VALUES ACROSS EACH OF THE THREE BLOCKS. R = MEAN REWARD VALUE ACROSS TRIALS. BLOCK 3 RESULTS PROVIDE DATA FOR *TERMINAL ACCURACY*. VALUE IN PARENTHESIS IS STANDARD DEVIATION. ΔR = DIFFERENCE IN MEAN REWARD VALUE FROM PREVIOUS BLOCK (*LEARNING RATE*). P = P-VALUE RESULT AFTER BONFERRONI CORRECTION. * DENOTES STATISTICAL SIGNIFICANT DIFFERENCE.

| Condition | Metric | Block 1 Trials 2-9 | Block 2 Trials 10-17 | Block 3 Trials 18-25 |
|---|---|---|---|---|
| DOT + Greedy | R (SD) | -0.107 (0.329) | 0.179 (0.560) | 0.473 (0.421) |
| | ΔR | - | 0.286 | 0.295 |
| | p | - | .06 (1&2) | **.006\*** (2&3) **<.001\*** (1&3) |
| DOT + ε-greedy | R (SD) | -0.196 (0.299) | 0.188 (0.546) | 0.304 (0.482) |
| | ΔR | - | 0.384 | 0.116 |
| | p | - | **.003\*** (1&2) | .657 (2&3) **<.001\*** (1&3) |
| Non-DOT + Greedy | R (SD) | 0.000 (0.311) | 0.107 (0.437) | 0.170 (0.527) |
| | ΔR | - | 0.107 | 0.062 |
| | p | - | .536 (1&2) | 1.000 (2&3) .418 (1&3) |
| Non-DOT + ε-greedy | R (SD) | -0.036 (0.400) | 0.152 (0.410) | 0.259 (0.497) |
| | ΔR | - | 0.187 | 0.107 |
| | p | - | .144 (1&2) | 1.000 (2&3) .077 (1&3) |

### A. Mutual Learning (Mean Reward Evaluations)

To assess *terminal accuracy,* we found no statistically significant interaction between the effects of Human Training Type and Robot Reinforcement Learning Type on mean reward ($F(3, 108) = .126$, $p = .726$, partial $\eta^2 = .005$). We found a statistically significant main effect of **Human Training Type** (DOT vs. Non-DOT) on the mean reward attained in the final eight trials ($F(3, 108) = .488$, $p = .017$, partial $\eta^2 = .194$). No statistically significant main effect was found for **Robot Reinforcement Learning Type** (greedy vs. ε-greedy) on mean reward in the final eight trials ($F(3, 108) = .084$, $p = .775$, partial $\eta^2 = .003$).

In order to assess *learning rate*, we then conducted a one-way ANOVA with repeated measures to understand whether there was a significant difference in the mean reward attained over successive blocks (of eight trials). Details of the mean reward and changes in mean reward per block can be seen in Table II.

**DOT+greedy:** Mauchly's Test of Sphericity indicated that the assumption of sphericity was met ($p = 0.102$, $\chi^2(2) = 4.573$). We found a statistically significant difference in mean reward changes between all three blocks of trials ($p < 0.001$, $F(2, 81) = 17.682$, partial $\eta^2 = 0.396$). Analysis with Bonferroni adjustments found statistically significant differences between blocks 1 and 3, and blocks 2 and 3 (Table II).

**Non-DOT+greedy:** Mauchly's Test of Sphericity indicated that the assumption of sphericity was met ($p = 0.104$, $\chi^2(2) = 4.532$). We did not find a statistically significant difference in mean reward changes ($p = 0.238$, $F(2, 81) = 1.45$, partial $\eta^2 = 0.052$). Analysis with Bonferroni adjustments did not find any statistically significant differences between blocks (Table II).

TABLE III. MAIN THEMES AND SUB-THEMES FROM THEMATIC ANALYSIS

| Main Theme | Sub Theme | n |
|---|---|---|
| Interface Flaws | Difficulty handling the buttons (4)<br>Similar sounding words (3)<br>Hard to understand what was/was not a reward (2) | 7 |
| Anthropomorphism | Feeling guilty (towards Reachy) when underperforming (1)<br>Feeling pleased (towards Reachy) when performing well (1) | 2 |
| Negative Affective Experiences of Task | Confusing (11)<br>Frustrating (8)<br>Difficult (6)<br>Feelings of underperforming (4)<br>Not logical (2) | 31 |
| Neutral Affective Experiences of Task | Neither fun nor boring (2) | 2 |
| Positive Affective Experiences of Task | Fun (8)<br>Interesting (4)<br>Feeling of achievement (3) | 15 |

**DOT+ε-greedy:** Mauchly's Test of Sphericity indicated that the assumption of sphericity was met ($p = 0.738$, $\chi^2(2) = 0.607$). We found a statistically significant difference in mean reward changes between all three blocks of trials ($p < 0.001$, $F(2, 81) = 13.656$, partial $\eta^2 = 0.336$). Analysis with Bonferroni adjustments found statistically significant differences between blocks 1 and 2, and 1 and 3.

**Non-DOT+ε-greedy:** Mauchly's Test of Sphericity indicated that the assumption of sphericity was met ($p = 0.18$, $\chi^2(2) = 3.43$). We found a statistically significant difference in mean reward changes between all three blocks of trials ($p = 0.034$ $F(2, 81) = 3.618$, partial $\eta^2 = 0.118$). However, analysis with Bonferroni adjustments did not find statistically significant differences between blocks (Table II).

In summary, our results found statistically significant improvements in the mean reward attained for *terminal accuracy* by Reachy in DOT conditions (Human Training Type). We found statistically significant improved learning rate in DOT conditions and also the Non-DOT condition with ε-greedy Robot Reinforcement Learning Type.

*B. Participant Interview Results*

Responses provided by the participants were analysed using the thematic analysis methodology outlined by Braun and Clarke [30]. Our results found five main themes: interface flaws; anthropomorphic feelings towards Reachy, positive affect negative affect, and neutral affect with respect to the learning task. Table III below shows the overall results of our thematic analysis. We discuss this in more detail in Section V.

## V. DISCUSSION

Our results showed that the differential affective expressions (outcomes) that Reachy expressed had a statistically significant effect on mutual learning between humans and robots in a language acquisition task as measured in terms of terminal accuracy. Specifically, conditions where Reachy's response used differential outcomes (DOT) (for training the human interactors) resulted in larger mean reward values compared to when Reachy used non-differential outcomes (Non-DOT). These findings support our first hypothesis (H1) and lend additional weight to the preliminary work that we have built on [9].

Our results show that differential outcomes training - implemented here as different types of gestures performed by a robot - offers advantages to mutual learning between a human and a robot, compared to non-differential outcomes training. Exploiting the implicit (reward-related) learning route in the brain, hypothesized to be activated through differential outcomes associations [18], [19] to offer greater potential for such mutual learning than using the route linked to 'retrospective learning' (hypothesized to be exploited during non-differential outcomes training) whereby stimulus-response 'task rules' are learned in order to obtain rewarding outcomes.

Although our results did not find any statistically significant effects of Robot Reinforcement Learning Type (greedy vs. ε-greedy) on mutual learning between human and robot, we found that the exploration-exploitation strategy (ε-greedy) resulted in improved learning performance over time, regardless of the type of expressions that Reachy was performing (i.e. differential or non-differential outcomes). For the exploitation strategy (greedy), statistically significant differences were only observed with differential outcome training. These findings therefore support our second hypothesis (H2) that an exploration-exploitation policy selection, more commonly used in Reinforcement Leaning agents, can provide improved performance over time. This is in spite of the fact that, where Reachy changes 'preference' (through a sub-optimal exploratory action selection (i.e. a babble), the human participant is still able to learn. We consider this setting to be more realistic in terms of human-infant interactions, and more conducive to long-term human-robot interactions where changes in preferences in order to find optimal mutually selected actions/states is desirable. Moreover, in dynamic environments, exploratory behaviour allows for lifelong mutual learning. We therefore see in our results the potential for further development of SAR using mutual learning for long-term human-robot interactive partnerships.

Post-hoc analysis of our thematic analysis (Table III) also revealed that participants who spoke positively about the task tended to report better performance on the learning tasks. For example, when asked about the experiment, P3 described it as *"It was exciting. A fun little experiment"* (see Supplementary Materials). This participant reported a mean reward value of 0.75 (DOT+Greedy), 0.5 (DOT+ε-greedy), 0.75 (Non-DOT+Greedy), and 0.75 (Non-DOT+ ε-greedy) in the final block of trials: performing above the mean in all conditions (Table II). The same is true for P20, who stated that the tasks were *"cool, interesting"*, and also reported a mean reward value above the mean in three of four conditions. While it is difficult to establish any relationship between positive affective attitudes to the mutual learning task and performance with our current data, these representative case studies provide some preliminary insight into a possible relationship between the two, which we aim to evaluate more deeply in the future. This may also align with other work in human-robot interaction that finds similar effects between enjoyment, affective engagement, and task performance [31], [32]. Moreover, long-term motivation to persist in human-robot interactions, e.g. for companionship, education, interventions, benefits not just from strong learning performance but also from subjectively evaluated positive experience.

Looking at the results from the final eight trials (block 3), we do not find a significant difference between the two different types of Q-learning policies that the robot employs (exploitation, or exploration-exploitation) (Table II). We propose that this may be due to the small epsilon value that we used in our present study (0.1). Though this value is a standard epsilon value in reinforcement learning, a 0.1 probability means Reachy would select a random action approximately 2 or 3 times over 24 trials. It may therefore be the case that Reachy had few opportunities to "explore" alternative actions in these conditions. Though we did not investigate the number of times Reachy used exploration strategies, or the distribution of the exploration over the course of each session, we plan this analysis for future work. We also plan to test different probabilities for exploration (different epsilon values) as well as dynamic exploration probabilities (i.e. via a decaying epsilon) in future work. Nevertheless, the statistically significant improvements on mutual learning that we found in this condition (Table II) suggest some value in allowing for probabilistic exploration of actions, which can be built on either by refining the current Q-learning approach, or leveraging related approaches, such as active inference [33]. The fact, however, that epsilon-greedy based Q-learning did not prohibit learning, in spite of the potential confusion it might cause (with occasional changing robot preferences), hints at its potential for use as a long-term learning approach for human-robot interactions in settings where interaction is intended to endure over weeks to months (as for cognitive interventions).

The above-mentioned results may also have been affected by the within-subject design, where each participant interacted with all four conditions. Despite instructions that Reachy "resets" at the start of each condition and that the human actor should also forget all previous word-object associations, there was still a risk that participants' previous associations would spill over into new conditions. It is possible that the study design contributed to the mentions of the paradigm being "frustrating" (e.g. P16, P18, P20) or "confusing" (e.g. P14, P23). This may, in part, be mitigated in the future by using a between-subjects study design. Order control of conditions, however, should have statistically mitigated such effects to a reasonable degree.

Several participants (e.g. P5, P9, P16) also mentioned that the method of interaction with the Unity simulator was not intuitive. The current set up required the pressing of one of three keys ("c", "d", or "b" to move the cookie, drink, and teddy, respectively). Despite the fact that these keys were marked on the keyboard, some participants stated that they pressed the wrong key and, therefore, selected the wrong option by accident. This therefore affected their performance (in terms of mean reward values), but also contributed to feelings of confusion and frustration with the setup.

Additionally, some participants (e.g. P2, P4, P20) reported that they found Reachy's babbles—specifically, "bada" and "paba" —too similar, resulting in additional uncertainty about what Reachy was trying to communicate, thereby affecting performance. We speculate that this may have been further confounded by the synthesized babbling used in this study (see Supplementary Materials). These issues could be overcome through an alternative selection of babbling phrases, as well as the use of human-recorded (as opposed to synthesized) audio files.

Considering our potential areas of application for this research, such as education and healthy ageing (Section I), the negative affective attitudes to the current experiment set up must be taken seriously by researchers extending or building on the work presented here, regardless of the positive results that were found on mutual learning. For ethical as well as practical reasons, the cause(s) of these types of responses must be identified and mitigated as much as possible as part of the ongoing research process. Here, we offered some of the insights and limitations from our present study to inform more rigorous study design in the future. We hypothesise that addressing these issues would result in larger improvements in both mutual learning, as well as enjoyment, of this human-robot co-learning task.

## VI. CONCLUSION

In this paper, we have evaluated the use of affective-linguistic communication, in combination with differential outcomes training, on a human-robot mutual learning task. We evaluated mutual learning through a human-robot collaborative task. The robot was required to learn how best to communicate its internal (homeostatically-controlled) "needs" through various "babbling" sounds, whilst the human partner (a "caregiver") needed to learn which object the robot was asking for. Our results found statistically significant advantages to mutual learning through the use of differential outcomes training, and an increased rate of (mutual) learning when the robot leveraged exploration-exploitation policy selection. Our findings demonstrate the potential for mutual human-robot learning through differential outcomes exploiting internal and expressed affective-linguistic states to be used in applications that require long-term interactive use.

ETHICAL STATEMENT

All volunteering participants whose data were used in the work presented in this paper provided informed written consent to participate in the study. The study was conducted in accordance with the ethical standards of the institutional and national research committee and with the 1964 Helsinki declaration and its later amendments.